\documentclass{article}

     \PassOptionsToPackage{numbers, compress}{natbib}
\usepackage[final]{svrhm_2019}

\usepackage[pdftex]{graphicx}
\usepackage{amsmath,bm}

\usepackage[utf8]{inputenc} % allow utf-8 input
\usepackage[T1]{fontenc}    % use 8-bit T1 fonts
\usepackage{hyperref}       % hyperlinks
\usepackage{url}            % simple URL typesetting
\usepackage{booktabs}       % professional-quality tables
\usepackage{amsfonts}       % blackboard math symbols
\usepackage{nicefrac}       % compact symbols for 1/2, etc.
\usepackage{microtype}      % microtypography

\usepackage{chngcntr}       % for counterwithin before TexLive 2018

\newcommand{\argmax}{\mathop{\rm arg~max}\limits}

\title{
Biologically-Motivated Deep Learning Method
using Hierarchical Competitive Learning
}

\author{%
  Takashi Shinozaki \\
  Center for Information and Neural Networks (CiNet), \\
  National Institute of Information and Communications Technology (NICT), \\
  1-4 Yamadaoka, Suita, Osaka 565-0871, Japan \\
  Graduate School of Information Science and Technology, Osaka University, \\
  1-5 Yamadaoka, Suita, Osaka 565-0871, Japan \\
  \texttt{tshino@nict.go.jp} \\
}

\begin{document}

\maketitle

\begin{abstract}
This study proposes a novel biologically-motivated learning method 
for deep convolutional neural networks (CNNs).
The combination of CNNs and back
propagation (BP) learning is the most powerful method
in recent machine learning regimes.
However, it requires large labeled data
for training, and this requirement
can occasionally become a barrier for real world applications.
To address this problem and utilize unlabeled data,
I propose to introduce unsupervised competitive learning
which only requires forward propagating signals 
as a pre-training method for CNNs. 
The method was evaluated by image discrimination tasks using 
MNIST, CIFAR-10, and ImageNet datasets, 
and it achieved a state-of-the-art performance as a biologically-motivated method 
in the ImageNet experiment. 
The results suggested that the method enables higher-level learning
representations solely from forward propagating signals 
without a backward error signal for the learning of convolutional layers.
The proposed method could be useful for a variety of poorly labeled data, for example,
time series or medical data.
\end{abstract}

\section{Introduction}
Convolutional neural networks (CNNs) \cite{Krizhevsky2012} 
and back propagation (BP) learning \cite{Rumelhart1986}
are the most powerful combination of recent machine learning methods.
However, when these learning methods are likened to information processing 
in the brain, several objections mainly directed toward 
the BP of error information are raised \cite{Bartunov2018}. 
Conversely, 
the brain is a more powerful learning machine 
than any current deep learning systems, 
which  simultaneously realizes 
the possibility of scaling computations to a very large network 
and strong semi-supervised learning.
Therefore, 
it can be said that clarifying the operational principles of the brain
is important for realizing a superior learning machine.

In recent years, 
biologically-motivated methods have predominantly been studied 
in cases where learning can be performed by estimating BP errors 
obtained from other feedback signals without the actual usage of BP errors
\cite{tp1,tp2,tp3,dtp,fa1,fa2}. 
However, 
because these methods also represent supervised learning using labeled data, 
 in which considerable labeled data are required for learning, 
brain-like computing cannot be realized.
To solve this problem, a version of unsupervised learning 
that does not use any BP information is required.

Numerous studies on 
unsupervised learning of visual representation have been conducted. 
It was initially studied using analytical methods.
Some previous studies applied independent component analysis to obtain
visual bases from natural scene images \cite{Olshausen1997,Hyvarinen2001}.
This method was subsequently expanded to a neural network (NN)-like regime \cite{Le2012}.
Because these methods were vastly different from conventional NNs
that were trained using BP learning, the unification of neural and
nonneural mechanisms has been challenging.

By contrast, similar to an NN-like architecture, the
Boltzmann machine and its families were applied
for pre-training to initialize weight parameters
\cite{Hinton2002,Hinton2006}. 
However, the structure and dynamics of these two NNs are
completely different. Thus, there is some inefficiency in the conversion
from the obtained learning representation in the Boltzmann machine
to a target feedforward NN.

Autoencoder \cite{Bengio2007} might address the conversion problem;
it is a variation of the feedforward NN, and
the weight parameter is wholly applicable to a traditional feedforward NN.
However, 
CNNs require  additional processes 
because of the non-linearity of the deconvolution processes.
Thus, generator-based unsupervised learning is used for the smooth
transfer of representation learning \cite{Radford2016}.
Goroshin \cite{Goroshin2015} used a sophisticated approach that transfers
the visual bases directly into the sub-network of the target CNN
using static object information in movies. 
However, these networks still rely on back-propagated signals, 
and learning methods that do not depend on 
back-propagated signals are nonetheless considered necessary.

Competitive learning is a method of unsupervised learning 
that learns by a simple sparse mechanism called "winner takes all" (WTA).
Competitive learning does not use any BP information, 
and instead simply uses local feedforward information. 
It has been used in networks termed
self-organizing maps (SOMs) \cite{Kohonen1982}
and Neocognitron \cite{Fukushima1980},
and recently
it has attracted attention as a plausible biological learning method 
\cite{Shinozaki2016,Shinozaki2017,Krotov2019}.
However, the conventional method is not always compatible with CNNs, which is the basic structure of the present deep neural network (DNN).
In this study, I propose a novel method to apply competitive learning to a multilayer CNN.

One of the most important mechanisms supporting the recent development of CNNs is
a rectified linear unit (ReLU) activation function,
which realizes efficient sparsification through a combination with BP learning. 
However, it does not work sufficiently without BP signals.
Therefore, 
I propose a method that uses WTA as an activation function, 
which enables sparse interlayer propagation without BP information.
Autoencoder using the WTA mechanism as an activation function is mainly studied.
\cite{Makhzani2014,Srivastava2013}.
In this study, I combine the WTA activation function
and competitive learning, which also uses WTA dynamics, 
and realize unsupervised learning without BP signals in a DNN.

The proposed method was evaluated by an image discrimination task
and demonstrated a drastic acceleration in the initial learning speed. 
Further, it achieved a state-of-the-art performance as a biologically-motivated method for the top-5 test error categories in the ImageNet experiment. 
The results suggested that 
this method obtains higher-level learning representation from unlabeled data 
and is usable for the following gradient-based fine tuning.

%----------------------------------------------------------------------
\section{Learning Methods}
\subsection{Competitive Learning}
The proposed method was based on traditional competitive learning 
\cite{Rumelhart1985,Grossberg1987},
which is used in Neocognitron \cite{Fukushima1980} and 
SOMs \cite{Kohonen1982}. 
I used the simplest competitive learning method, i.e., 
the WTA algorithm, 
and did not use any position information over the filter axis. 
The WTA algorithm is only applied to  the weight parameter update process 
and has no effect on the feedforward signal. 
The WTA algorithm was performed for the input vector of the $l$-th layer, 
which is described as $\bm{u}_{l}=\bm{W}_{l} \bm{z}_{l-1}$, 
where $\bm{z}_{l-1}$ is the output vector of the previous layer 
and $\bm{W}_{l}$ is the connection matrix. 
If the activation function is a monotone increasing function, 
the unit with the maximum input value is the one with the maximum output value. 
I termed it the "winner" unit and performed a weight update 
of the competitive learning only for that unit. 
Therefore, the weight gradient of the $i$-th unit of the $l$-th layer was 
described as follows: 
\begin{equation}
  \Delta \bm{w}_{l,i}=
  \begin{cases}
    - \rho \bm{z}_{l-1}, & \text{if } i=\argmax_{k} u_{l,k} \\
    0, & \text{otherwise},
  \end{cases}
\end{equation}
where $\rho$ is the learning coefficient of competitive learning, 
and was set to $0.5 \times 10^{-3}$. 
The weight vector was normalized by L2-norm at every update 
to stabilize the competitive learning \cite{Kohonen1982}: 
\begin{eqnarray}
\bm{w'}_{l,i} = (\bm{w}_{l,i} + \Delta \bm{w}_{l,i})
/ ||\bm{w}_{l,i} + \Delta \bm{w}_{l,i}||.
\label{eq:norm}
\end{eqnarray}

We also introduces the conscience factor proposed by DeSieno \cite{DeSieno1988}
for the winner processing 
to improve the learning efficiency for large networks. 
It adjusts the balance of winning ratio among units, 
preventing only some units becoming dominant.
It gives the unit with the initial noise pattern a chance to win 
and enables equal learning for all units.
The conscience factor in this study was described as follows:
\begin{equation}
  b_i = C(1/N - p_i).
\end{equation}
where $C$ is the constant of conscience, and I empirically
determined it to be $5.0$ for the whole network.
$N$ is the number of units of the target layer, and
$p_i$ is the probability of winning for the $i$-th unit in the minibatch.
The final version of the weight gradient of competitive learning
was described as follows: 
\begin{equation}
  \Delta \bm{w}_{l,i}=
  \begin{cases}
    - \rho \bm{z}_{l-1}, & \text{if } i=\argmax_{k} (u_{l,k} + b_k) \\
    0, & \text{otherwise},
  \end{cases}
\end{equation}

Competitive learning is treated as unsupervised pre-training. 
First, it extracts the basis 
(e.g., Gabor patches for natural scene images and harmonics for audio)
from unlabeled input data. 
Then, gradient-based learning is applied only 
to the last fully-connected layer as a fine-tuning 
using the learning representation obtained through competitive learning. 
It is important to note that the competitive learning 
is applied to the entire network at once, which is fundamentally different from the conventional autoencoders 
which mostly behave in a layer-wise manner.

%--------------------------------------------------
\subsection{WTA as an activation function}
In recent neural networks, sparse dynamics are essential for DNNs, 
and ReLU is their key. 
As sparse dynamics are strongly controlled by bias factors, 
which are modulated by BP learning, 
 different form of sparse dynamics were required for the proposed method. 
WTA is one of the most simple activation functions for sparseness. 
Thus, instead of ReLU, I employed WTA again 
as an activation function for convolutional layers
. 
The WTA algorithm for each feature map functioned similar to 
a functional column in the biological brain. 
Each location in each channel has a corresponding winner. 
For example, an output with 5 x 5 pixels in a single channel has 25 winners, 
and is a bundle of 25 one-hot vectors. 
%

%----------------------------------------------------------------------
\section{Experiments}
To investigate the proposed method's adaptation to a difficult dataset, 
I employed the ImageNet dataset \cite{imagenet} for verification. 
Many potential biological models have not been tested 
on a dataset possessing this level of complexity. 
The results were compared with the few previous studies that tackled 
the problem \cite{Bartunov2018}. 
I also evaluated  two simpler image datasets: 
MNIST \cite{mnist} and CIFAR-10\cite{cifar}. 
Their results have been described in the appendix. 

I employed Alexnet \cite{Krizhevsky2012} as the baseline, 
and modified it for the application of 
the proposed hierarchical competitive learning. 
Fig.\ref{fig:struct}(c) shows the network structure for 
the ImageNet task. 
I reduced the number of layers with information
for the same spatial resolution,
and increased the number of channels of spatial frequency 
in respective convolutional layers. 
Eventually,
the network constituted three convolutional 
and one fully-connected layers. 
The first convolutional layer comprised 256 filters with 3 x 3 pixels. 
The second and third layers possessed 
an inception\cite{Szegedy2015}-like structure
to avoid obtaining only low frequency filters, 
and the layers consisted of 1024 filters with 3 x 3 and 5 x 5 pixels. 
The convolutional layers were accompanied by 
the WTA activation function and 4 x 4 maxpooling with a 2-pixel stride. 
The fully-connected layer had 1000 units
corresponding to the number of labels of ImageNet. 

The size of the mini-batch was 8 for competitive learning and 
64 for the subsequent fine-tuning. 
The data augmentation comprised horizontal flips and
random crops from five positions (i.e., center, upper right, upper left, 
lower right, and lower left). 
The number of iterations for the pre-training using
competitive learning and fine-tuning using gradient-based learning were 
150,000 and 60,000, respectively.
All learning processes employed a
conventional stochastic gradient descent (SGD) method for the weight update 
using a learning coefficient of 0.01, 
and averaged cross-entropy was employed for the loss function. 
The learning rate of the fine tuning was reduced to one tenth 
every 20,000 iterations. 
I employed LeCun normal \cite{LeCun1998b} for weight initialization.

All codes were implemented using Python and 
Chainer deep learning framework (v.4.5.0) \cite{Tokui2015} 
with GPU support. 
All experiments were performed on an NVIDIA DGX-1 with Tesla P100, 
and CUDA (v.9.0) and cuDNN (v.7.1.4) libraries were used.
My code is available at \url{https://github.com/t-shinozaki/convcp/}.

Tab.\ref{tab:imagenet_err} shows the results of the experiment. 
The proposed method outperformed previous biologically-motivated methods, 
and achieved a state-of-the-art performance for both top-1 and top-5 test error categories.

\begin{table}[htb]
  \begin{center}
    \begin{tabular}{lrr}
      Method     &  Top-1    &  Top-5 \\
      \hline
      Ours                        &  \textbf{87.72} & \textbf{73.84} \\
      DTP, Parallel   \cite{Bartunov2018} &  98.34    &  94.56 \\
      SDTP, Parallel  \cite{Bartunov2018} &  99.28    &  97.15 \\
      FA              \cite{Bartunov2018} &  93.08    &  82.54 \\
      \\
      BP, ConvNet \cite{Bartunov2018} &  
      \underline{\textbf{63.93}} &  \underline{\textbf{40.17}} \\
    \end{tabular}
  \end{center}
  \caption{Test errors on ImageNet.}
  \label{tab:imagenet_err}
\end{table}

%----------------------------------------------------------------------
\section{Discussion}

In this study, I proposed a novel unsupervised pre-training method 
for CNNs using multi-layer competitive learning 
with a WTA activation function and conscience factor.
In the experiment, the competitive learning algorithm obtained high-level features
from unlabeled data 
and achieved state-of-the-art results 
as a biologically-motivated method in the ImageNet experiment. 

With the proposed method,
a larger network was needed to realize learning at the same level
as a conventional method.
It is generally considered that
the features obtained by conventional BP learning are selected and reduced 
according to the target task, 
whereas those obtained by competitive learning are all-inclusive 
over total input signals. 
The ability to utilize more feature quantities is expected 
to increase the resistance to adversarial attacks 
\cite{Goodfellow2015,Athalye2018}; 
However, this resistance is accompanied by an increase in the network size.
Although these characteristics could provide flexibility for several tasks,
pruning might be necessary for the efficient inference of a specific task.

%--------------
An advantage of the proposed method is 
that it can acquire more diverse learning representations 
compared with conventional BP learning. 
Moreover, this diverse learning representation provides strong adaptability 
to various types of data, which implies that the proposed method is suitable for increasing the number of filters. 
By contrast, the representation learning of conventional CNNs is relatively weak
particularly in their early layers, 
and they experience difficulty in increasing the number of filters 
near the input side because of the degradation of BP signals.
As a result, BP learning sometimes requires repetitive structures 
with or without residual connections and 
acquires the same spatial level filter set across multiple layers. 
This might be one of the reasons
why recent DNNs possess extremely deep structures. 
The proposed method could enlarge the number of filters in CNNs 
and enable broad and effective information processing in each layer. 

Finally, because the implementation of the proposed method enables 
seamless switching between
unsupervised learning using competitive learning and
supervised learning using gradient-based learning,
the method could also be useful for the mixture condition of the two 
learning methods termed as semi-supervised learning. 
Furthermore, the proposed method is also expected to be 
applied to few-shot learning scenarios because it can effectively utilize 
unlabeled data.
However, further studies are required.

%----------------------------------------------------------------------
\subsubsection*{Acknowledgments}
This work was supported by JST ERATO Grant Number JPMJER1801, Japan.

%------------------------------------------------------------
% APPENDIX
%------------------------------------------------------------
\newpage
%\vspace{0.5cm}
\appendix
\counterwithin{figure}{section}
\counterwithin{table}{section}

\section{Network Structures}

\begin{figure}[htb]
  \begin{center}
    \includegraphics[width=0.95\linewidth]{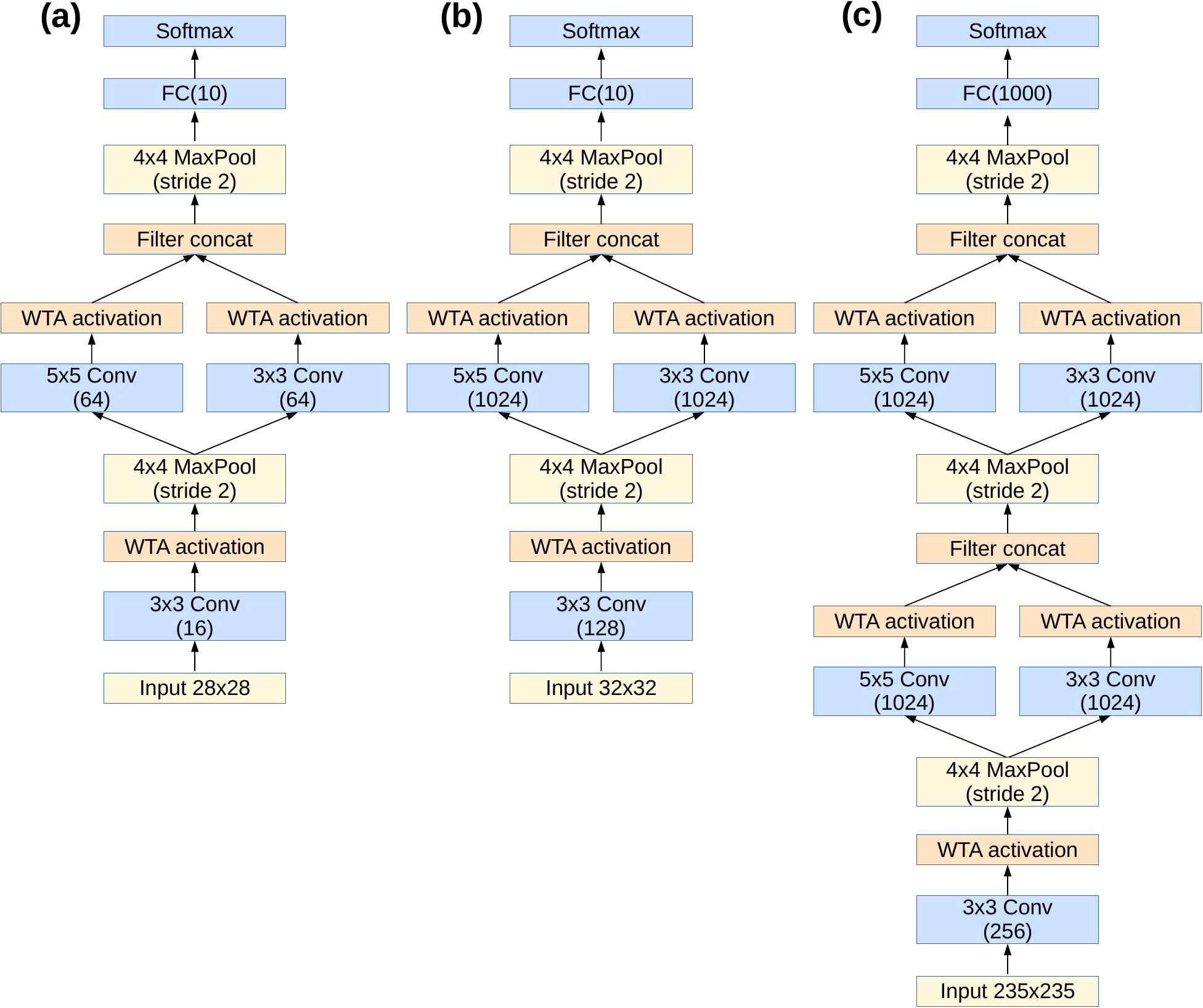} \\
    \caption{Network structures for (a) MNIST, (b) CIFAR, and 
      (c) ImageNet experiments.}
    \label{fig:struct}
  \end{center}
\end{figure}

%----------------------------------------
\section{Other Datasets}
I performed image discrimination tasks with MNIST \cite{mnist}, 
and CIFAR-10\cite{cifar} datasets in addition to the ImageNet task. 
A  LeNet5-like neural network \cite{LeCun1989} was employed as the baseline 
and the proposed hierarchical competitive learning on was applied to it. 
The network consisted of two or three convolutional layers 
and one fully-connected layer. 
Each convolutional layer was followed by an activation function, 
WTA (for the test) or ReLU (for baseline), and max-pooling. 
The detailed structure of the network for each task 
is shown in Fig.\ref{fig:struct}. 
I compared test errors of image discrimination results.

\subsection{MNIST}
I employed an LeNet5-like network for the test networks using only two convolutional layers. 
Fig.\ref{fig:struct}(a) shows the detailed structure of the network. 
The baseline network consisted of two convolutional 
and two fully-connected layers. 
The convolutional layers consisted of 
25 and 50 filters with 5 x 5 pixels 
and were accompanied by ReLU and 2x2 maxpooling. 
The fully-connected layers consisted of 100 and 10 units.

Both pre-training and fine-tuning employed 
the training dataset with 50,000 samples, 
and the validation process used the test dataset with 10,000 samples. 
The size of a mini-batch was 100, 
and the averaged cross-entropy was employed for the loss function. 
All learning processes undertook a
conventional SGD method for the weight update, 
and the learning coefficienct was 0.01. 
I did not employing weight decay or momentum. 
The number of iterations for the pre-training using competitive learning 
and fine-tuning using gradient-based learning are 15,000 and 3,000, respectively.
It should be noted that image augmentation was not employed.

\begin{figure}[htb]
  \raisebox{3.4cm}{\bf (a)} \hspace{-0.10cm}
  \raisebox{1.8cm}{
    \includegraphics[width=0.13\linewidth]{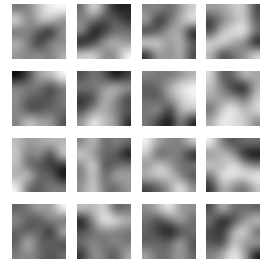}}
  \hspace{0.2cm}
  \raisebox{3.4cm}{\bf (b)} \hspace{-0.10cm}
  \includegraphics[width=0.25\linewidth]{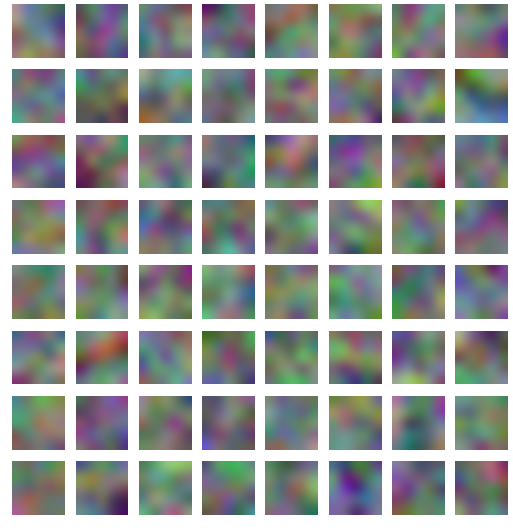}
  \hspace{0.2cm}
  \raisebox{3.4cm}{\bf (c)} \hspace{-0.10cm}
  \raisebox{1.8cm}{
    \includegraphics[width=0.13\linewidth]{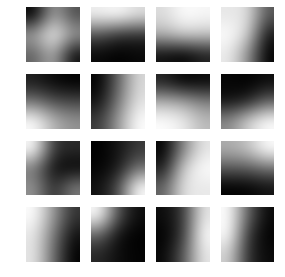}}
  \hspace{0.2cm}
  \raisebox{3.4cm}{\bf (d)} \hspace{-0.10cm}
  \includegraphics[width=0.25\linewidth]{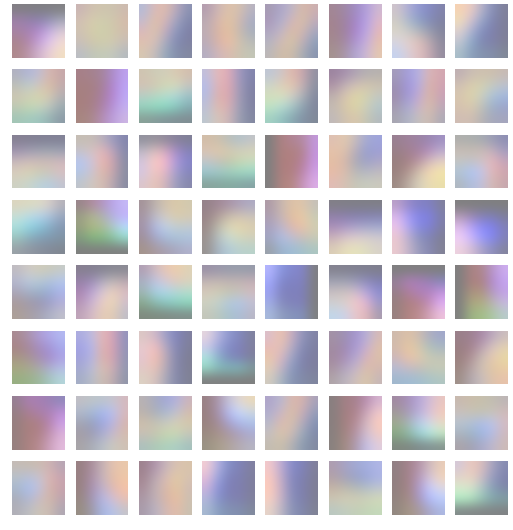}
  \caption{Obtained filters in MNIST experiments of 
    (a) first and (b) second convolutional layers 
    learned by conventional BP learning. 
    (c,d) those by competitive learning.}
  \label{fig:mnist_filt}
\end{figure}

Fig.\ref{fig:mnist_filt} shows the obtained filters 
in the first and second convolutional layers.
The color of the filter was obtained by 
dividing the entire filter set into three bins 
corresponding to red, green, and blue respectively.

Fig.\ref{fig:mnist_filt}(a,b) were learned using 
only conventional BP learning, and looks like noise patterns. 
By contrast, 
the filter set learned through competitive learning obtained 
clear spatial structures (Fig.\ref{fig:mnist_filt}(c,d)).
The results show that competitive learning performs 
stronger representation learning than BP learning.
% ????

\begin{figure}[htb]
  \begin{minipage}[c]{0.48\hsize}
    \begin{center}
      \includegraphics[width=0.60\linewidth]{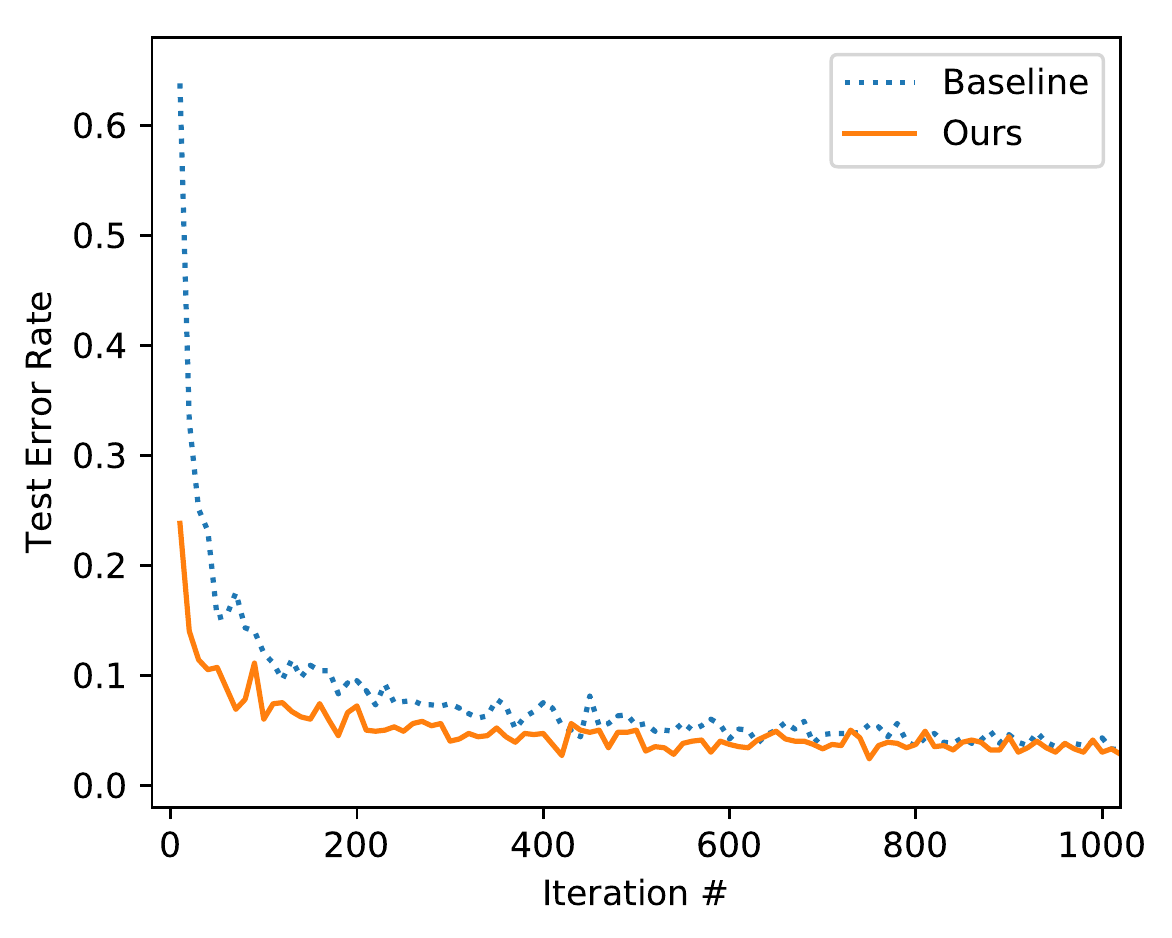}
      \caption{Transitions of test errors during fine-tuning 
        in MNIST experiments.}
      \label{fig:mnist_trans}
    \end{center}
  \end{minipage}
  \hfil
  \begin{minipage}[c]{0.48\hsize}
    \makeatletter
    \def\@captype{table}
    \makeatother
    \begin{center}
      \begin{tabular}{lrr}
        Method     & MNIST \\
        \hline
        Baseline   & \underline{\textbf{0.90}} \\  % {1.82} \\
        Ours       & 1.79 \\  % 2.24 \\
        \\
        DTP, Parallel, LC   \cite{Bartunov2018} &  \textbf{1.52} \\
        SDTP, Parallel, LC  \cite{Bartunov2018} &  1.98 \\
        FA, LC              \cite{Bartunov2018} &  1.85 \\
      \end{tabular}
    \end{center}
    \caption{Test errors on MNIST.}
    \label{tab:mnist_err}
  \end{minipage}
\end{figure}

Fig.\ref{fig:mnist_trans} shows the transition of test errors during the fine-tuning process. 
The proposed method converged much faster than the baseline and achieved almost the same accuracy. 
Tab.\ref{tab:mnist_err} shows the resultant test error rates for respective methods. 
Because  image augmentation was not employed, the result of the baseline was not outstanding.
However, the result from the proposed method is comparable to other biologically motivated learning methods.

\vspace{0.5cm}

%----------------------------------------
\subsection{CIFAR}
I employed LeNet5-like neural network with two convolutional layers.
The network was almost identical to that of MNIST, 
but possessed an extension of the number of filters.
Fig.\ref{fig:struct}(b) shows the detailed structure of the network. 
The baseline network consisted of three convolutional 
and two fully-connected layers. 
The convolutional layers consisted of 
32, 32, and 64 filters with 5 x 5 pixels 
and were accompanied by ReLU and 2 x 2 maxpooling. 
The fully-connected layers comprised 4096 and 10 units.

Both pre-training and fine-tuning used 
the training dataset with 50,000 samples, 
and the validation process employed the test dataset with 10,000 samples. 
The size of a mini-batch was 100, 
and the averaged cross-entropy was employed for the loss function. 

All learning processes employed the
normal SGD method for the weight update, 
and the learning coefficient was 0.01. 
I did not use either weight decay or momentum. 
The number of iterations for the pre-training
using competitive learning and fine-tuning using gradient-based learning
are 75,000 and 15,000, respectively. Image augmentation was not employed. 

\begin{figure}[htb]
  \raisebox{2.8cm}{\bf (a)} \hspace{-0.10cm}
  \raisebox{1.3cm}{
  \includegraphics[width=0.12\linewidth]{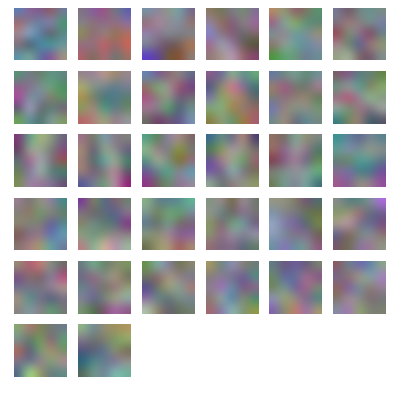}}
  \raisebox{2.8cm}{\bf (b)} \hspace{-0.10cm}
  \includegraphics[width=0.233\linewidth]{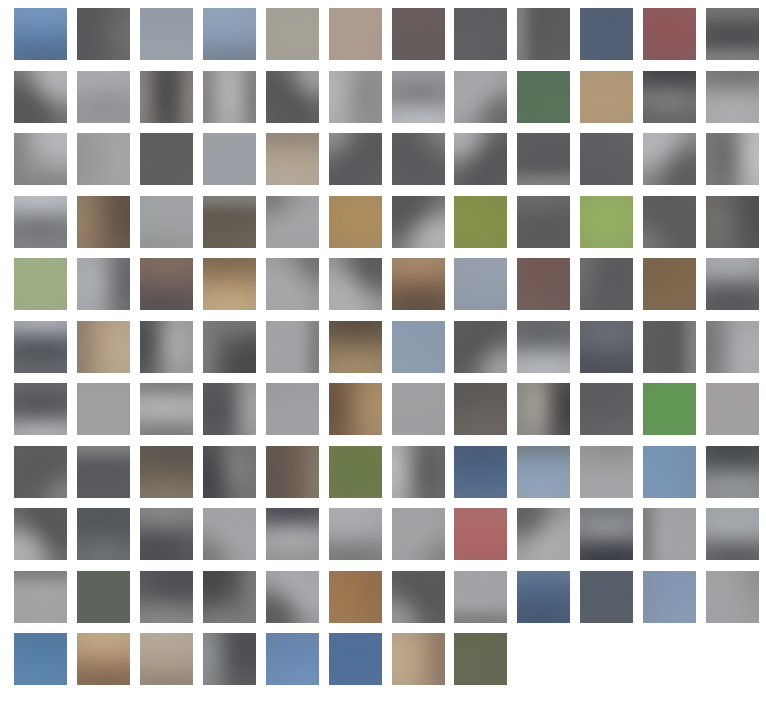}
  \raisebox{2.8cm}{\bf (c)} \hspace{-0.10cm}
  \includegraphics[width=0.233\linewidth]{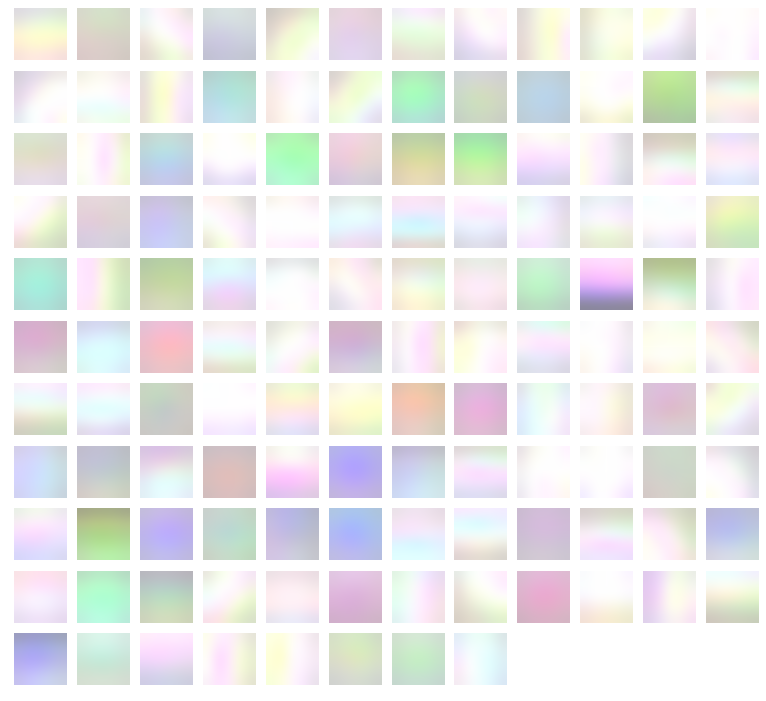}
  \raisebox{2.8cm}{\bf (d)} \hspace{-0.10cm}
  \includegraphics[width=0.233\linewidth]{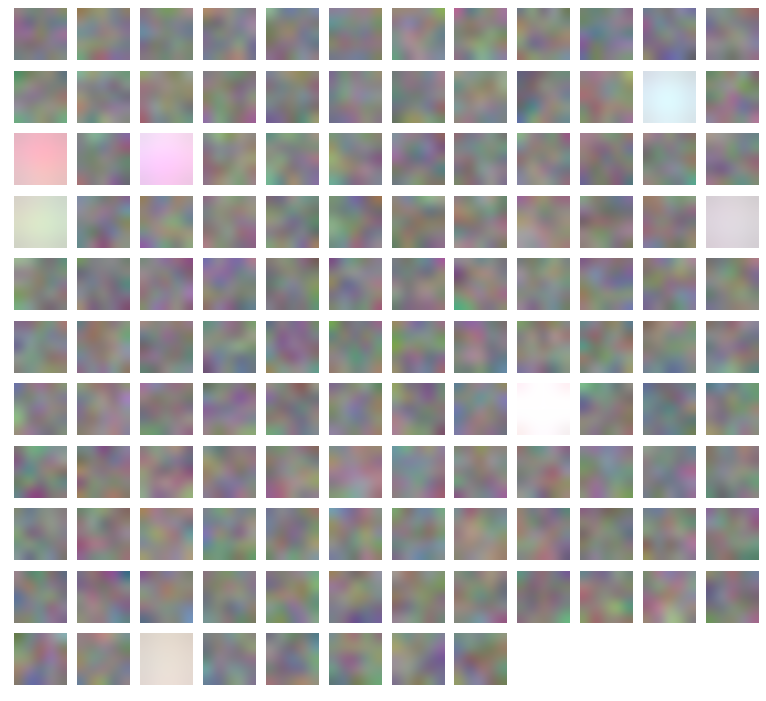}
  \caption{Obtained filters in CIFAR experiments. 
    Filters in the first layer with 5x5 pixel size 
    learned by (a) conventional back propagation learning or 
    (b) competitive learning;
    (c) Filters in the second layer learned by competitive learning employing the conscience factor; 
    (d) (c) without the conscience factor.}
  \label{fig:cifar_filt}
\end{figure}

Fig.\ref{fig:cifar_filt} shows the obtained filter sets
in conv1 (b) and conv2 (c) learned through competitive learning. 
Although BP learning could not achieve clear spatial structure at the filter sets 
(Fig.\ref{fig:cifar_filt}(a)), 
the proposed method can obtain a basis with clear spatial structure 
for natural images.
However, filters obtained by competitive learning were dominated by 
low spatial frequency because the dynamics were strongly dependent 
on their frequency of occurrence. 
Hence, inception\cite{Szegedy2015}-like structure was introduced to acquire features with
higher spatial frequencies using small filters. 

Fig.\ref{fig:cifar_filt}(d) shows the filter set with the second 
convolutional layer learned through competitive learning 
without the conscience factor (CF) \cite{DeSieno1988}.
Only a few filters could obtain a clear spatial structure, 
while the others were obtained as noise patterns.
As this tendency was more pronounced in higher (output side) layers, 
applying CF is essential when using competitive learning 
for DNNs. 

\begin{figure}[htb]
  \begin{minipage}[c]{0.48\hsize}
    \begin{center}
      \begin{center}
        \includegraphics[width=0.60\linewidth]{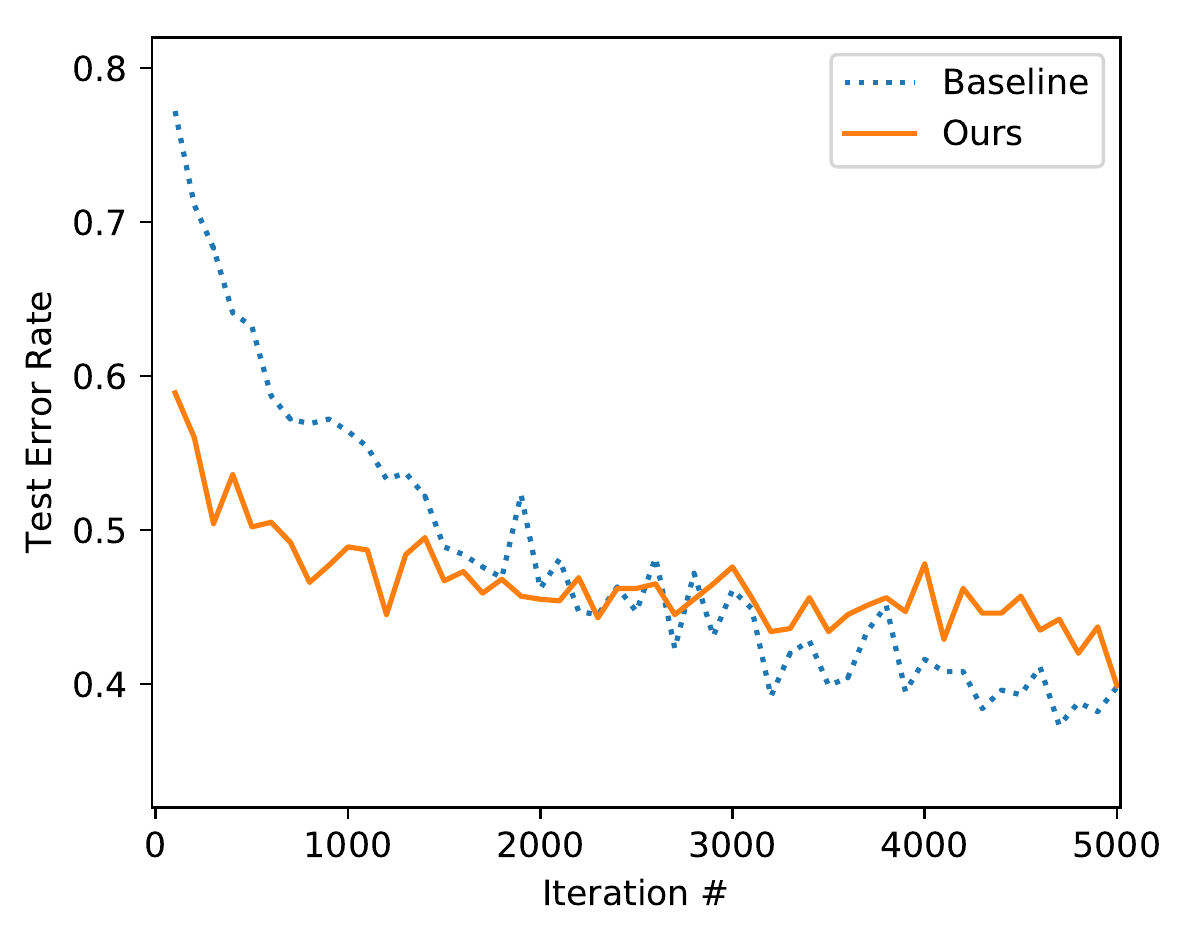}
      \end{center}
      \caption{Transitions of test errors during fine-tuning 
        in CIFAR experiments.}
      \label{fig:cifar_trans}
    \end{center}
  \end{minipage}
  \hfil
  \begin{minipage}[c]{0.48\hsize}
    \makeatletter
    \def\@captype{table}
    \makeatother
    \begin{center}
      \begin{tabular}{lrr}
        Method     & CIFAR \\
        \hline
        Baseline   & \textbf{37.74} \\
        Ours       & 39.31 \\
        \\
        DTP, Parallel,    \cite{Bartunov2018} &  39.47 \\
        SDTP, Parallel,   \cite{Bartunov2018} &  46.63 \\
        FA,               \cite{Bartunov2018} &  \underline{\textbf{37.44}} \\
        % \\
      \end{tabular}
    \end{center}
    \caption{Test errors on CIFAR.}
    \label{tab:cifar_err}
  \end{minipage}
\end{figure}

Fig.\ref{fig:cifar_trans} shows the transitions of the test errors during fine-tuning.
Competitive learning 
drastically accelerated the learning speed during the initial iterations
(before 2000 iterations), and the test errors were comparable 
with the baseline.

Tab.\ref{tab:cifar_err} % \textit{upper} 
shows the comparison of the final test errors among several methods. 
The proposed method achieved a comparable score 
with respect to biologically motivated learning and BP. 

\vspace{0.5cm}

\end{document}